%% file: paper.tex
\def\year{2021}\relax
\documentclass[letterpaper]{article} 
\usepackage{aaai21}  
\usepackage{times}  
\usepackage{helvet} 
\usepackage{courier}  
\usepackage[hyphens]{url}  
\usepackage{graphicx} 
\usepackage{graphicx}  
\usepackage{natbib}  
\usepackage{caption} 
\usepackage{subfig}
\usepackage{amsmath,amssymb,amsfonts,amsthm}
\usepackage{multirow}
\usepackage{xcolor}
\usepackage[ruled, vlined, linesnumbered]{algorithm2e}

\usepackage{comment}
\usepackage{cite}
\usepackage{subfig}

\newcommand{\R}{\mathcal{R}}
\newcommand{\fml}[1]{\mathcal{#1}}

\usepackage{caption}
\DeclareCaptionLabelFormat{algnonumber}{Algorithm}
\usepackage{todonotes}
\usepackage{listings}

\newcommand{\argmin}{\operatornamewithlimits{argmin}}

\newboolean{includeMemo}
\setboolean{includeMemo}{true} 

\newcommand{\memo}[1]{\ifthenelse{\boolean{includeMemo}}{\todo[inline,caption={},color=green!20!]{#1}}}
\newcommand{\memob}[1]{\ifthenelse{\boolean{includeMemo}}{\todo[inline,caption={},color=blue!20!]{#1}}}

\newtheorem{definition}{{\bf Definition}}

\newtheorem{lemma}{{\bf Lemma}}

\newtheorem{proposition}{{\bf Proposition}}

\newcommand{\bitemize}{\begin{list}{$\bullet$}{\topsep=1pt \parsep=0pt \itemsep=1pt \leftmargin=1em }} 
\newcommand{\eitemize}{\end{list}}
\newcommand{\beitemize}{\begin{list}{$\bullet$}{\topsep=1.5pt \parsep=0pt \itemsep=1pt \leftmargin=1em }} 
\newcommand{\enitemize}{\end{list}}

\definecolor{apricot}{rgb}{0.98, 0.81, 0.69}

\newcommand{\M}{\mathcal{M}}

\newcommand{\C}{\mathcal{C}}
\newcommand{\gmcs}{getMCS}
\newcommand{\gmus}{getMUS}

\title{On Exploiting Hitting Sets for Model Reconciliation\footnote{This paper has been published in AAAI 2021.}}
\author{
    Stylianos Loukas Vasileiou,\textsuperscript{\rm 1}
     Alessandro Previti,  \textsuperscript{\rm 2}
        William Yeoh \textsuperscript{\rm 1} \\
    }
\affiliations{
    \textsuperscript{\rm 1} Washington University in St. Louis \\
    \textsuperscript{\rm 2} Ericsson Research \\
    \{v.stylianos, wyeoh\}@wustl.edu,
     alessandro.previti@ericsson.com
  }

\begin{document}

\maketitle

\begin{abstract}
In human-aware planning, a planning agent may need to provide an explanation to a human user on why its plan is optimal. A popular approach to do this is called \emph{model reconciliation}, where the agent tries to reconcile the differences in its model and the human's model such that the plan is also optimal in the human's model. In this paper, we present a logic-based framework for model reconciliation that extends beyond the realm of planning. More specifically, given a knowledge base $KB_1$ entailing a formula $\varphi$ and a second knowledge base $KB_2$ not entailing it, model reconciliation seeks an explanation, in the form of a cardinality-minimal subset of $KB_1$, whose integration into $KB_2$ makes the entailment possible. Our approach, based on ideas originating in the context of analysis of inconsistencies, exploits the existing hitting set duality between \emph{minimal correction sets} (MCSes) and \emph{minimal unsatisfiable sets} (MUSes) in order to identify an appropriate explanation. However, differently from those works targeting inconsistent formulas, which assume a single knowledge base, MCSes and MUSes are computed over two distinct knowledge bases. We conclude our paper with an empirical evaluation of the newly introduced approach on planning instances, where we show how it outperforms an existing state-of-the-art solver, and generic non-planning instances from recent SAT competitions, for which no other solver exists. 
\end{abstract}

\input{introduction}

\input{preliminaries}

\input{framework}

\input{computing}

\input{experiments}

\input{related}

\input{conclusions}

\section*{Acknowledgments}

This research is partially supported by NSF grant 1812619. The views and conclusions contained in this document are those of the authors and should not be interpreted as representing the official policies, either expressed or implied, of the sponsoring organizations, agencies, or the U.S. government.

\bibliography{references}

\end{document}

%% file: introduction.tex
\section{Introduction}

With increasing proliferation and integration of AI systems in our daily life, there is a surge of interest in \emph{explainable AI}, which includes the development of AI systems whose actions can be easily understood by humans. Driven by this goal, \emph{machine learning} (ML) researchers have begun to classify commonly used ML algorithms according to different dimensions of explainability~\cite{Guidotti:2018:SME:3271482.3236009}; improved the explainability of existing ML algorithms~\cite{alvarez2018towards,petkovic2018improving}; as well as proposed new ML algorithms that trade off accuracy for increasing explainability~\cite{dong2017improving,gilpin2018explaining}.

In contrast, researchers in the \emph{automated planning} community have mostly taken a complementary approach. While there is some work on adapting planning algorithms to find easily explainable plans\footnote{Also called \emph{explicable} plans in the planning literature.} (i.e.,~plans that are easily understood and accepted by a human user)~\cite{zhang2017plan}, most work has focused on the \emph{explanation generation problem} (i.e.,~the problem of identifying explanations of plans found by planning agents that when presented to users, will allow them to understand and accept the proposed plan)~\cite{kambhampati1990classification,langley2016explainable}. Within this context, there is a popular theme that has recently emerged called \textit{model reconciliation}~\cite{chakraborti2017plan}. Researchers in this area have looked at how an agent can explain its decisions to a human user who might have a different understanding of the same planning problem. These explanations bring the human's model closer to the agent's model by transferring the minimum number of updates from the agent's model to the human's model. However, a common thread across most of these works is that they, not surprisingly, employ mostly automated planning approaches. 

In this paper, we approach the model reconciliation problem from a different perspective -- one based on  \emph{knowledge representation and reasoning} (KR). We propose a novel general logic-based framework for model reconciliation, where given a knowledge base $KB_a$ (of an agent) that entails a formula $\varphi$ and a knowledge base $KB_h$ (of a human user) that does not entail $\varphi$, the goal is to identify a subset of $KB_a$ such that when it used to update $KB_{h}$, then $KB_{h}$ entails $\varphi$. More specifically, we present a novel algorithm that exploits the hitting set duality of \emph{minimal correction sets} (MCSes) and \emph{minimal unsatisfiable sets} (MUSes) for computing minimum explanations with respect to two knowledge bases. 
MUSes and MCSes have also been studied by \citet{rei87} under the name of conflicts and diagnoses, respectively. \citet{KleerMR92} have related MUSes (conflicts) and MCSes (diagnoses) to prime implicates and prime implicants.
Further, we implement the proposed algorithm in propositional logic and evaluate its performance against the current state-of-the-art~\cite{chakraborti2017plan} on classical planning problems as well as present results on some general instances from recent SAT competitions.
Although presented in the context of propositional logic, the algorithm can be applied to any type of constraint system for which the satisfiability of subsets can be decided.
Our empirical results demonstrate that our approach significantly outperforms the current state of the art when the explanations are long or when the difference between the agent's and human's models is large, and that it is efficient and feasible for problems beyond planning.


%% file: preliminaries.tex
\section{Preliminaries}

\subsection{Classical Planning}

A \emph{classical planning} problem is a tuple $ \Pi := \langle D, I, G\rangle$, which consists of the domain $D = \langle F, A \rangle$ -- where $F$ is a finite set of fluents representing the world states ($s\in F$) and $A$ a set of actions -- and the initial and goal states $I, G \subseteq F$. An action $a$ is a tuple $\langle pre_{a}, eff^{\pm}_{a}\rangle$, where $pre_{a}$ are the preconditions of $a$ -- conditions that must hold for the action to be applied; and $eff^{\pm}_{a}$ are the addition ($+$) and deletion ($-$) effects of $a$ -- conditions that must hold after the action is applied. The solution to a planning problem $\Pi$ is a plan $\pi = \langle a_{1}, \ldots , a_{n} \rangle$ such that $\delta _{\Pi}(I, \pi) = G$, where $\delta_{\Pi}(\cdot)$ is the transition function of problem $\Pi$. The cost of a plan $\pi$ is given by $C(\pi, \Pi) = |\pi|$. Finally, a cost-minimal plan $\pi^{*} = \argmin_{\pi \in \{\pi' \mid \delta_{\Pi}(I, \pi') = G\}}  C(\pi, \Pi)$ is called the optimal plan.

\subsection{Explainable AI Planning}

\emph{Explainable AI Planning} (XAIP), as introduced by \citet{ijcai19}, couples the model of the human user and the planning agent's own model into its deliberative process. Therefore, when there exist differences between those two models such that the agent's optimal plan diverges from the user's expectations, the agent attempts a \emph{model reconciliation} process. In this process, the agent provides an explanation that can be used to update the user's model such that the agent's plan is also optimal in the updated user's model. 

More formally, a \emph{Model Reconciliation Problem} (MRP)~\cite{chakraborti2017plan} is defined by the tuple $\Psi = \langle \varphi, \pi \rangle $, where $\varphi = \langle M^a, M^a_h \rangle$ is a tuple of the agent's model $M^a = \langle D^a, I^a, G^a \rangle$ and the agent's approximation of the user's model $M^a_h = \langle D^a_h, I^a_h, G^a_h \rangle$, and $\pi$ is the optimal plan in $M^a$. A solution to an MRP is an explanation $\epsilon$ such that when it is used to update the user's model $M^a_h$ to $\widehat{M}^{a, \epsilon}_h$, the plan $\pi$ is optimal in both the agent's model $M^a$ and the updated user's model $\widehat{M}^{a, \epsilon}_h$. The goal is to find a shortest explanation.

\subsection{Propositional Logic}
\label{logic}

In this section, we provide the basic definitions used throughout the paper. Additional standard definitions are assumed~\cite{sat-handbook}. 
Although the new algorithm can be applied to any constraint
system for which the satisfiability of constraint subsets can be checked, the focus of this paper is on propositional logic.

A formula in \emph{conjunctive normal form} (CNF) is a conjunction of clauses, where each clause is a disjunction of literals. A literal is either a Boolean variable or
its negation. For convenience, and when it is clear from the context, we might refer to formulas as sets of clauses and clauses as
sets for literals. An \emph{interpretation} $I : V \rightarrow \{0, 1\}$ is a mapping from the set of variables $V$ to $\{0, 1\}$. A formula is \emph{satisfiable} if there exists an interpretation that satisfies it. A satisfying interpretation is referred to as a \emph{model}.
A formula is \emph{unsatisfiable} or \emph{inconsistent} when no model exists. In what follows, we assume the knowledge base $KB$ and all the formulas are always expressed in CNF.
This is not a restrictive requirement, since any propositional formula can be transformed into a CNF. Moreover, unless stated otherwise, $KB$ will be assumed to be consistent.


\begin{definition}
[Entailment]
	A formula $\varphi$ is logically entailed by $KB$, denoted by $KB \vDash \varphi$, if every model of $KB$ is also a model of $\varphi$.
    $KB \vDash \varphi$ iff $KB \land \lnot \varphi$ is unsatisfiable.
\end{definition}


\begin{definition}
[Minimal Unsatisfiable Set (MUS)]
	Given an inconsistent $KB$, a subset $\M \subseteq KB$ is an MUS if $\M$ is unsatisfiable and $\forall \M' \subset \M$, $\M'$ is satisfiable. 
\end{definition}

By definition, every unsatisfiable $KB$ contains at least one MUS. 

\begin{definition}
[Minimal Correction Set (MCS)]
	Given an inconsistent $KB$, a subset $\C$ of $KB$ is an MCS if $KB \setminus \C$ is satisfiable and $\forall \C' \subset \C$ we have that $KB \setminus \C'$ is unsatisfiable.
\end{definition}
\begin{definition}
	A set of clauses $P$ is a partial MUS of an inconsistent $KB$ if it exists at least one MUS $M \subseteq KB$ such that $P \subseteq M$. 
\end{definition}

Partial MUSes appear when in a inconsistent KB a subset of clauses is set as hard.
MUSes and MCSes are related by the concept of \emph{minimal hitting set}.

\begin{definition}
[Hitting Set]
	Given a collection $\Gamma$ of sets from a universe $U$, a hitting set for $\Gamma$ is a set $H \subseteq U$ such that $\forall S \in \Gamma$, $H \cap S \neq \emptyset$.
\end{definition}

A hitting set is \emph{minimal} if none of its subsets is a hitting set. The relationship between MUSes and MCSes is discussed by~\citet{kas-jar08} and \citet{liffiton-pmm16} and it was firstly presented by~\citet{rei87}, where MUSes and MCSes are referred to as (minimal) conflicts and diagnoses, respectively.

\begin{proposition}\label{prop:duality}
	A subset $\M$ ($\C$) of an inconsistent $KB$ is an MUS (MCS) iff it is a minimal hitting set of the collection of all MCSes (MUSes) of $KB$. 
\end{proposition}

It follows from the above proposition that a cardinality minimal MUS (MCS) is a minimal hitting set. \emph{Cardinality minimal MUS} are referred
to as SMUS, whereas a \emph{cardinality minimal MCS} corresponds to the complement of a MaxSAT solution~\cite{li2009maxsat}. We also refer to a cardinality minimal set as a minimum or smallest set.

\begin{lemma}\label{lem:smus}
	Given a subset $\fml{H}$ of all the MCSes of $KB$, a hitting set is an SMUS if:
	\begin{enumerate}
		\item It is a minimum hitting set $h$ of $\fml{H}$, and 
		\item The subformula induced by $h$ is inconsistent. 
	\end{enumerate}
\end{lemma}

\noindent See the work by~\citet{ignatiev-plm15} for a proof.
Proposition \ref{prop:duality} and Lemma \ref{lem:smus} naturally extend to the case of partial MUS.
Note that when some clauses are set as hard in an inconsistent knowledge base, the set of all MCSes is a subset of the one where all the clauses in the knowledge base are set as soft. In this case, every minimal hitting set on the set of all MCSes is a partial MUS.

\begin{definition}
[Support]
	Given a $KB$ s.t. $KB \vDash \varphi$, a support for $\varphi$ is a subset $\epsilon \subseteq KB$ such that $\epsilon \vDash \varphi$ and $\forall \epsilon' \subset \epsilon$ we have
	$\epsilon' \nvDash \varphi$.
\end{definition}

In what follows, given a formula $F$, we will write $F^*$ with $* \in \{s, h\}$ to denote a set of clauses that will be treated as \emph{soft} and \emph{hard}, respectively.
Intuitively, the hard clauses are those clauses that will not be removed by the minimization procedure.

MUSes and supports are related by the following:

\begin{proposition}\label{prop:support-mus}
	A consistent set of clauses $\epsilon$ is a support of $\varphi$ ($\epsilon \vDash \varphi$) iff $\epsilon$ is a partial MUS of $\epsilon \land \lnot \varphi$.
\end{proposition}

In what follows, we provide a definition of a \emph{logic-based} model reconciliation problem~\cite{vas}:

\begin{definition}
[Model Reconciliation]
	Given two knowledge bases $KB_a$ and $KB_h$ of the agent providing an explanation and the human receiving the explanation, respectively, such that $KB_a \vDash \varphi$ and $KB_h \nvDash \varphi$, the goal of model reconciliation
	is to find a support\footnote{Note that we use the term ``support'' to mean ``explanation'' in the traditional sense in this context.} $\epsilon \subseteq KB_a \land KB_h$ such that $KB_h \land \epsilon \vDash \varphi$. 
\label{lmrp}
\end{definition}

We refer to the set of clauses $\epsilon \setminus KB_h$ as the  \emph{update} of the knowledge base $KB_h$.
In this paper, we focus on the more specific task of computing a support $\epsilon$ such that $\epsilon \setminus KB_h$ is an update of minimum size. 

\begin{definition}
[Partial Support]
	A subset $\epsilon_p$ is a partial support if there exists at least one support $\epsilon$ such
	that $\epsilon_p \subseteq \epsilon$.
\end{definition}
\noindent 
Given a formula $\epsilon \land \lnot \varphi$, $\epsilon_{p}$ is thus a subset of the partial MUS $\epsilon$. An update is a partial support.

\subsubsection{Encoding Classical Planning Problems as SAT:}

A classical planning problem can be encoded as a SAT problem~\cite{kautz1992planning,kautz1996encoding}. The basic idea is the following: Given a planning problem $P$, find a solution for $P$ of length $n$ by creating a propositional formula that represents the initial state, goal state, and the action dynamics for $n$ time steps. This is referred to as the \emph{bounded planning problem} $(P,n)$, and we define the formula for $(P,n)$ such that: \emph{Any} model of the formula represents a solution to $(P,n)$ and if $(P,n)$ has a solution, then the formula is satisfiable.

We encode $(P,n)$ as a formula $\varphi$ involving one variable for each action $a \in A$ at each timestep $0\leq i < n$ and one variable for each fluent $f \in F$ at each timestep  $0\leq i \leq n$. We denote the variable representing action $a$ in timestep $i$ using subscript $a_i$, and similarly for facts. The formula $\varphi$ is constructed such that $\langle a_0, a_1, \ldots , a_{n-1} \rangle$ is a solution for $(P,n)$ if and only if $\varphi$ can be satisfied in a way that makes the fluents $a_0, a_1, \ldots , a_{n-1}$ true. Finally, we can \emph{extract} a plan by finding a model that satisfies $\varphi$ (i.e.,~for all time steps  $0\leq i < n$, there will be exactly one action $a$ such that $a_i = $ \emph{True}). This could be easily done by using a satisfiability algorithm, such as the well-known DPLL algorithm~\cite{DPLL62}. 

%% file: framework.tex
\section{}

%% file: computing.tex
\section{Computing Explanations}


\citet{ignatiev-plm15} introduced an algorithm for computing a smallest MUS of an inconsistent knowledge base $KB$. Building on that approach, we introduce a new algorithm that computes a smallest support $\epsilon$ for a formula $\varphi$ that needs to be explained. 
The idea is to reduce the problem of computing a support of minimum size to the one of computing an SMUS over an inconsistent formula.
Notice that by definition, we have that $KB \vDash \varphi$ iff $KB \land \lnot \varphi$ is unsatisfiable.
Moreover, in Proposition \ref{prop:support-mus}, we have already
stated the relation between a support and an MUS. This suggests that, in order to extract a support, we just need to run an MUS solver over the formula
$KB_h^s \land \lnot \varphi^h$,\footnote{Recall that $KB_h^s$ denote that the clauses in the human's $KB_h$ are treated as \emph{soft} while $\varphi^h$ denote that the clauses in the formula $\varphi$ that needs to be explained are treated as \emph{hard}. Soft clauses may be removed by the MUS solver while hard clauses will not.} and then remove $\lnot \varphi$ from the returned MUS.
The duality relating MUSes and MCSes is a key aspect for the computation of an SMUS. In the next section, we will show how this duality can be exploited for the task of model reconciliation. We first start by reporting the algorithm for computing a smallest support for the case of a single knowledge base, and then we will illustrate how this approach can be revised for the case of two knowledge bases.
\begin{algorithm}[t]
\DontPrintSemicolon
	\KwIn{$KB, \varphi$ }
	\KwResult{A minimum size support $\epsilon$}
	$\fml{H}\gets\emptyset$\; \label{alg1:initH}
	\While{true}{
		\tcp{Compute a minimum hitting set}
		$seed\gets minHS(\fml{H})$\;\label{alg1:hit}
		$\epsilon \gets \{c_i \;|\;  i \in seed \}$\;\label{alg1:iseed}
		\uIf{not $SAT(\epsilon \land \lnot \varphi)$}{ \label{alg1:stcheck}
			\tcp{minimum size support}
			\Return $\epsilon$\;
		}
		\Else{
			$\C \gets \gmcs(seed, KB^s \land \lnot \varphi^h)$\; \label{alg1:mcs}
		}
		$\fml{H}\gets \fml{H}\cup \{ \C \}$\;
	}
	\caption{Basic algorithm for computing the smallest support (one KB)}
	\label{alg:sexp}
\end{algorithm}
\begin{algorithm*}[t]
\DontPrintSemicolon
	\KwIn{$KB_a, KB_h, \varphi$ }
	\KwResult{An explanation $\epsilon$ such that $\epsilon \setminus KB_h$ is of minimum size}
	$\R \gets\emptyset$\;\label{alg:bmdlr:rec}
	$KB_a^h \gets KB_a \cap KB_h$\tcp*{hard clauses}\label{alg:bmdlr:kbh}
	$KB_a^s \gets KB_a \setminus KB_a^h$\tcp*{soft clauses}\label{alg:bmdlr:kbs}
	\If{not $SAT(KB_h \land KB_{a})$\label{alg:bmdlr:sat-check-1}}{
		$\C \gets  \gmcs((KB_{h} \setminus KB_{a})^{s} \land KB_{a}^{h})$\label{alg:bmdlr:preprocessing-1}\;
		$KB_{h} = KB_{h} \setminus \C$\label{alg:bmdlr:preprocessing-2}\;
	}
	\While{true}{\label{alg:bmdlr:loop}
		$seed\gets minHS(\R)$\;\label{alg:bmdlr:mhs}
		$\epsilon_{p} \gets \{c_i \; | \; i \in seed \}$\tcp*{A partial support $\epsilon_p$ induced by the seed}
	    \uIf{not $SAT(KB_h \land \epsilon_p \land \lnot \varphi)$}{\label{alg:bmdlr:sat-check}
			$\epsilon \gets \gmus(KB_h^s \land \epsilon_{p}^h \land \lnot \varphi^h) \setminus \lnot \varphi$\;\label{alg:bmdlr:return}
			\Return $\epsilon$\;
		}
		\Else{
			$\C \gets \gmcs(seed, KB_h^h \land KB_a^s \land \lnot \varphi^{h})$\;\label{alg:bmdlr:mcs}
			$\R \gets \R \cup \{ \C \}$\;
		}
		
	}
	\caption{Model Reconciliation Algorithm}
	\label{alg:bmdlr}
\end{algorithm*}
Algorithm \ref{alg:sexp} is based on the algorithm for computing a smallest MUS originally presented by~\citet{ignatiev-plm15}.
$\fml{H}$ is a collection of sets, where each set corresponds to an MCS on $KB$. At the beginning, it is initialized with the empty set (line~\ref{alg1:initH}). 
Each MCS in $\fml{H}$ is represented as the set of the indexes of the clauses in it. $\fml{H}$ stores the MCSes computed so far.
At each step,  a minimum hitting set on $\fml{H}$ is computed (line \ref{alg1:hit}). On line \ref{alg1:iseed}, the
formula induced by the computed minimum hitting set is stored in $\epsilon$. Then, the formula $\epsilon \land \lnot \varphi$ is tested for satisfiability
(line \ref{alg1:stcheck}). If $\epsilon \land \lnot \varphi$ is unsatisfiable, then $\epsilon$ is a support of minimum size. The algorithm return $\epsilon$
and the procedure ends. If instead $\epsilon \land \lnot \varphi$ is satisfiable, then it means that $\epsilon \nvDash \varphi$ and the algorithm continues at
line \ref{alg1:mcs}.
The computation
of an MCS of this kind can be performed via standard MCS procedures~\cite{marques-silva-ijcai13}, using the set of clauses indexed by the \emph{seed} as
the starting formula to extend. Since the clauses in $\varphi$ are set to hard (line \ref{alg1:mcs}), the returned MCS $\C$ is guaranteed to be contained
in $KB$. Due to the hitting set duality relation, we will also have $\epsilon \subseteq KB$. Notice that the procedure $getMCS$ always reports a new MCS because, by construction, we have $seed \subseteq KB \setminus \C$. In fact, the $seed$ contains at least one clause for each previously computed
MCS and thus $seed \cap \C = \emptyset$ (i.e., at least one clause for each previously computed MCS is not in $\C$).

\subsection{Model Reconciliation}

In the previous section, we presented an approach for the computation of the smallest support for the case of a single knowledge base. 
Here, we show how this method can be further extended for the case of two knowledge bases, a task that we defined as \emph{model
reconciliation}. In what follows, we assume that
one is the knowledge base of an agent $KB_a$ and the other is the one of a human $KB_h$. The task we are targeting is to
find a support $\epsilon \subseteq KB_a \land KB_h$ such that $KB_h \land \epsilon \vDash \varphi$ and $\epsilon \setminus KB_h$ is of minimum size. $\R$ is the formula we use to store the MCSes,
which acts as a mediator between $KB_a$ and $KB_h$.
Intuitively, the idea is to compute MCSes over $KB_a$, add them to $\R$, and then test the minimum hitting sets over $KB_h$.

Algorithm \ref{alg:bmdlr} summarizes the main steps of this new approach. At the beginning of the algorithm,
$\R$ is empty (line~\ref{alg:bmdlr:rec}). Lines~\ref{alg:bmdlr:kbh} and~\ref{alg:bmdlr:kbs} are used to specify which clauses of $KB_a$
will be treated as hard and soft, respectively. We then check if $KB_h \land KB_a$ is inconsistent (line~\ref{alg:bmdlr:sat-check-1}).
This is important in order to avoid the possibility of finding subsets $\epsilon$ that explains why $KB_h \land KB_a$ is inconsistent instead of the target support.
In case $KB_h \land KB_a$ is inconsistent, we preprocess $KB_h$ by removing from $KB_h \setminus KB_a$ a minimal set of clauses causing the conflict (i.e., an MCS) (lines~\ref{alg:bmdlr:preprocessing-1}-\ref{alg:bmdlr:preprocessing-2}). The reconciliation procedure
starts on line \ref{alg:bmdlr:loop}. The algorithm proceeds iteratively by computing a minimum hitting set on $\R$ and then testing for satisfiability
the induced subformula $\epsilon_p$. $\epsilon_p$ is an under approximation of the final partial support.
The test checks whether updating $KB_h$ with $\epsilon_p$ is sufficient for entailing $\varphi$. If the formula $KB_h \land \epsilon_p \land \lnot \varphi$
is unsatisfiable, then an MUS containing a subset of $KB_h$, $\epsilon_p$, and $\lnot \varphi$ is returned, and the set of clauses $\lnot \varphi$ is removed from it.
The result, from Proposition \ref{prop:support-mus}, is an $\epsilon = M \land \epsilon_p$, with $M \subseteq KB_h$, such that $\epsilon \vDash \varphi$. Otherwise, the algorithm continues on line \ref{alg:bmdlr:mcs},
where a new MCS is computed and added to $\R$. 
%
%
The algorithm is complete in the sense that eventually a support $\epsilon$ such that $\epsilon \setminus KB_h$ is of minimum size will be returned. This can be easily verified by observing
that every time $KB_h \land \epsilon_p \land \lnot \varphi$ is satisfiable, a new MCS is computed. Eventually, all the MCSes will be computed and, from
Propositions~\ref{prop:duality} and~\ref{prop:support-mus}, it follows that a minimum hitting set on the collection of all MCSes corresponds to the
smallest update. 
Deciding whether there exists a support of size less or equal to k is $\Sigma_{2}^{p}$-complete and extracting a smallest support is in $FP^{\Sigma_{2}^{p}}$. This follows directly from the complexity of deciding and computing an SMUS on which Algorithm \ref{alg:bmdlr} is based \cite{ignatiev-plm15}.

\begin{table*}[t]
 \resizebox{1\linewidth}{!} {
	\begin{tabular}{ | l l l | }
		\hline
		 & $KB_a =  \; \stackrel{C_1}{(a \lor b)} \land \stackrel{C_2}{(\lnot b \lor c)} \land \lnot \stackrel{C_3}{c} \land \stackrel{C_4}{(\lnot b \lor d)} \land \lnot \stackrel{C_5}{d}$ & \multirow{2}{*}{\Bigg\} We have that $KB_a \vDash a$ and $KB_h \nvDash a$} \\
		 & $KB_h = \; \stackrel{D_1}{\lnot c} \land \stackrel{D_2}{f}$ & \\[1mm]
		\hline
		1 & $\fml{R} \gets \emptyset$ &  \\
		2 & $KB_a^h \gets KB_a \cap KB_h = \{C_3\}$ & \\
		3 & $KB_a^s \gets KB_a \setminus KB_a^h = \{C_1, C_2, C_4, C_5\}$ & \\
		4 & $seed \gets \emptyset$ & \# $minHS(\fml{R})$ \\
		5 & $\lnot c \land f \land \emptyset \nvDash a$ & \# $SAT(KB_h \land \epsilon_p \land \lnot \varphi)$ \\
		6 & $\fml{C} \gets \{C_2, C_4\}$ & \# MCS computed on $KB_h \land KB_a^s \land \lnot a$ starting with the seed $seed$ \\
		7 & $\fml{R} \gets \{\{C_2, C_4\}\}$ & \\
		8 & $seed \gets \{C_2\}$ & \# $minHS(\fml{R})$ \\
		9 & $\lnot c \land f \land (\lnot b \lor c) \nvDash a$ & \# $SAT(KB_h \land \epsilon_p \land \lnot \varphi)$ \\
		10 & $\fml{C} \gets \{C_1\}$ & \# MCS computed on $KB_h \land KB_a^s \land \lnot a$ starting with the seed $seed$ \\
		11 & $\fml{R} \gets \{\{C_2, C_4\}, \{C_1\}\}$ & \\
		12 & $seed \gets \{C_1, C_2\}$ & \# $minHS(\fml{R})$ \\
		13 & $\lnot c \land f \land (\lnot b \lor c) \land (a \lor b) \vDash a$ & \# $SAT(KB_h \land \epsilon_p \land \lnot \varphi)$ \\
		14 & $Return$ $\{C_1, C_2, D_1\}$ & \# MUS computed on $D_1 \land D_2 \land C_1 \land C_2 \land \lnot a \setminus \lnot a$ \\
		\hline
	\end{tabular}
	}
	\caption{Example of Algorithm \ref{alg:bmdlr}}
	\label{table:alg:bmdlr}
\end{table*}

Finally, notice that in some settings (e.g., planning), desired supports are required to be a subset of
$KB_a$. In this case, $KB_h$ is replaced with $KB_a^h$ on line~\ref{alg:bmdlr:sat-check}, $KB_h^s$ is replaced with $KB_a^h$ on line~\ref{alg:bmdlr:return}, and $KB_h^h$ is replaced with $KB_a^h$ on line  \ref{alg:bmdlr:mcs}.
Notice that, in this setting, the updated $KB_h \land \epsilon$ might be inconsistent. 
However, in our approach, this case is naturally resolved by the preprocessing step at line \ref{alg:bmdlr:preprocessing-2}.
In some settings, the update $\epsilon \setminus KB_h$ is expected to be a subset of a support $\epsilon' \vDash \varphi$ s.t. $\epsilon' \subseteq KB_a$. In this case the MCS returned at line \ref{alg:bmdlr:mcs} should be further shrinked in order to be an MCS of the sole $KB_a^{s} \land \lnot \varphi^h$.
Table \ref{table:alg:bmdlr} shows an example trace of Algorithm~\ref{alg:bmdlr}.

%% file: experiments.tex
\section{Experimental Evaluations}

We now present our experimental evaluation of Algorithm 2 for computing explanations on classical planning problems from the International Planning Competition as well as on some general problems from the SAT competition.\footnote{The code repository is: \url{https://github.com/vstylianos/aaai21}.}

\smallskip \noindent \textbf{Setup and Prototype Implementation:} We ran our experiments on a MacBook Pro machine comprising of an Intel Core i7 2.6GHz processor with 16GB of memory. The time limit was set to 1500s. Our implementation of Algorithm~2 is written in Python and integrates calls to SAT, MCS/MUS, and minimal hitting set oracles through the PySAT toolkit~\cite{imms-sat18}. 

\subsection{Classical Planning Instances}

\noindent \textbf{Explanations in Planning:} Following the MRP literature \cite{chakraborti2017plan}, we focus on the problem of explaining the \emph{optimality} of a plan to a human user. Recall that, in classical planning problems with uniform action costs, a plan $\pi^{*} $ is optimal if no shorter plan exists. Couched in terms of propositional logic, we say that a plan $\pi^{*}$ of length $n$ is optimal if, given a knowledge base $KB$ encoding the specific planning problem, no shorter plan exists in \emph{all} models of $KB$. Particularly, $\pi^{*}$ is optimal with respect to a knowledge base $KB$, if $KB \vDash \bigwedge_{t=0}^{n-1} \neg g_t$, where $g_t$ is the fact corresponding to the goal of the planning problem at time step $t$. However, in order to show that a plan $\pi^{*}$ is optimal, we would first need to show that $\pi^{*}$ is a feasible plan (i.e.,~the plan is sound and can be executed to achieve the goal). More specifically, we say that a plan $\pi^{*}$ is feasible with respect to a knowledge base $KB$ if $\pi^{*}$ and $g_{n}$ are true in at least one model of $KB$, where $g_{n}$ is the fact corresponding to the goal of the planning problem at the final time step $n$. Therefore, when combined with the fact that no plan of lengths $1$ to $n-1$ exists, then that plan $\pi^*$ must be an optimal plan.

Note that Algorithm 2 proposed in the previous section is agnostic to the underlying planning domain. However, it can be used to find explanations for MRP in explainable planning as follows: For checking if a plan is feasible, we impose the literals in $\pi^{*}$ and $g_{n}$ as assumptions in the SAT solver, and check if a model that satisfies them exists. If not, we add the missing actions as part of the explanation to $KB_{h}$. Finally, to find an explanation for the optimality of the plan, we use $\varphi = \bigwedge_{t=0}^{n-1} \neg g_t$ as the query and run Algorithm~2 with the changes to lines~\ref{alg:bmdlr:sat-check},~\ref{alg:bmdlr:return}, and~\ref{alg:bmdlr:mcs} as described in the last paragraph in the ``Model Reconciliation'' subsection. Then, the algorithm will return the smallest explanation that explains that the plan $\pi^*$ is both feasible and optimal. 

\smallskip \noindent \textbf{Experimental Scenarios}: We used the actual IPC instances as the model of the agent (i.e.,~$KB_a$) and tweaked that model by randomly removing parts (preconditions and effects) of the action model, and assigned it to be the model of the human user (i.e.,~$KB_h$). We used our own implementation of the encoding by \citet{kautz1996encoding} as the SAT encoding of the knowledge bases. We further make two important assumptions: (1)~First, we assume that $KB_a$ has the \emph{correct} and \emph{complete} encoding of the planning problem and only $KB_h$ can contain errors or omissions. (2)~We assume that $\pi^{*}$ is a feasible plan with respect to $KB_{a}$ and that $KB_{a} \vDash \bigwedge_{t=0}^{n-1} \neg g_t$. These assumptions are reasonable, especially when viewed from the perspective of the explaining agent, where explanations are based on the view (or model) of the specific problem~\cite{miller2018explanation}.

\begin{table*}[t]
  \centering
 \resizebox{1\linewidth}{!} {
\begin{tabular}{|cc||crr|crr|crr|crr|crr|cr|cr|cr|}
\hline
\multicolumn{2}{|c||}{\multirow{2}{*}{Prob.}} & \multicolumn{3}{c|}{Scenario 1} & \multicolumn{3}{c|}{Scenario 2} & \multicolumn{3}{c|}{Scenario 3} & \multicolumn{3}{c|}{Scenario 4} & \multicolumn{3}{c|}{Scenario 5} & \multicolumn{2}{c|}{Scenario 6} & \multicolumn{2}{c|}{Scenario 7} & \multicolumn{2}{c|}{Scenario 8}\\
\multicolumn{2}{|c||}{} & $|\epsilon|$ & {\sc cszk} &  {\sc alg2} & $|\epsilon|$ & {\sc cszk} & {\sc alg2} & $|\epsilon|$ & {\sc cszk} & {\sc alg2} & $|\epsilon|$ & {\sc cszk} & {\sc alg2} & $|\epsilon|$ & {\sc cszk} & {\sc alg2} & $|\epsilon|$ & {\sc alg2} & $|\epsilon|$ & {\sc alg2} & $|\epsilon|$  & {\sc alg2} \\
\hline
\hline
\multirow{4}{*}{\rotatebox{90}{\hspace{0.0em}\textsc{Blocks-}} \rotatebox{90}{\hspace{0.0em}\textsc{world}}} 
& 1 & 2 &  0.6 & \textbf{0.05s} & 1 & 0.2s & \textbf{0.05s} & 4 & 12.0s & \textbf{0.1s} & 8 & t/o& \textbf{0.2s} &5 & 129.5s & \textbf{0.2s} & 2 & \textbf{0.05s} & 6 & \textbf{0.3s}& 10 & \textbf{0.5s}
 \\
& 2 & 1 & 0.4s & \textbf{0.2s} & 1& \textbf{0.2s} & \textbf{0.2s} &4& 4.5s & \textbf{1.2s} & 6& t/o &\textbf{14.5s} & 5 &158.0s & \textbf{108.0s} & 4 & \textbf{0.2s} & 4 & \textbf{10.0s} &11& \textbf{118.0}
\\
& 3 & 3 & 1.5s &\textbf{0.5s} & 3 & 0.5s& \textbf{0.3s} & 6& 11.0s & \textbf{1.0s} & 7& t/o& \textbf{4.5s}&  8& 256.0s & \textbf{5.0s} & 1& \textbf{0.1s} & 9  & \textbf{5.0s} &12& \textbf{10.5s} 
\\
& 4 & 2& \textbf{1.0s} & 4.0s & 2& \textbf{0.2s} & 1.0s & 5& \textbf{9.5s} & 125.0s & 6& t/o & \textbf{136.5s}&  7&273.5s & \textbf{145.0s} & 12& \textbf{0.5s} & 8 & \textbf{150.0s} &--& t/o 
\\
\hline
\multirow{4}{*}{\rotatebox{90}{\hspace{0.0em}\textsc{Log-}} \rotatebox{90}{\hspace{0.0em}\textsc{istics}}} 
& 1 & 2& 1.0s & \textbf{0.3s} & 1& \textbf{0.2s} & 0.5s & 4& 47.0s & \textbf{0.5s} & 6 & t/o & \textbf{0.5s} & 5& 255.0s&\textbf{0.2s} &  3& \textbf{0.1s} & 7&\textbf{0.5s} & 8 &\textbf{1.0s} \\
& 2 & 2& \textbf{1.0s} & \textbf{1.0s} &  2& \textbf{0.2s} & 0.8s &  2& 1.0s & \textbf{0.6s} &  5 & t/o& \textbf{1.0s} & 6 &276.0s &\textbf{0.3s} &4 & \textbf{0.1s} & 6&\textbf{0.4s} & 7 & \textbf{1.0s} \\
& 3 & 2& 2.5s & \textbf{0.5s} &  3& 1.0s & \textbf{0.5s} & 3& 5.5s & \textbf{1.0s} &  6& t/o & \textbf{1.5s} & 6& 271.0s& \textbf{0.2s}& 6&\textbf{0.2s} &6& \textbf{0.2s}& 6 &\textbf{1.0s} \\
& 4 &4& 18.0s & \textbf{1.0s}& 2& \textbf{0.5s} & \textbf{0.5s} & 6& 117.0s & \textbf{2.0s} & 7& t/o & \textbf{2.5s} & 7 & 283.0s&\textbf{0.5s} & 7& \textbf{0.5s} &7& \textbf{0.5s}& 7 &\textbf{1.0s} \\
\hline
\multirow{4}{*}{\rotatebox{90}{\hspace{0.0em}\textsc{Rover}}} 
& 1 & 1 & 1.0s&  \textbf{0.2s} & 2&  \textbf{0.5s} & \textbf{0.5s} & 4 & 125.0s & \textbf{0.8s} & 7 & t/o & \textbf{1.0s}& 7&300.0s & \textbf{2.0s}& 3& \textbf{0.5s} &8& \textbf{2.5s}& 8& \textbf{3.0s} \\
& 2 & 3& 1.0s & \textbf{0.5s} &  1&  \textbf{0.5s} & \textbf{0.5s} &  2 & 1.0s & \textbf{0.5s} & 7 & t/o &  \textbf{2.0s}&  6 &311.0s &\textbf{3.0s} &5& \textbf{0.5s} &8 & \textbf{4.0s}& 9 &\textbf{4.0s}\\
& 3 & 1& \textbf{0.5s} & \textbf{0.5s} & 2& 1.0s & \textbf{0.5s} & 3& 55.0s & \textbf{1.0s} & 6 & t/o & \textbf{3.0s}&  6& 330.0s& \textbf{5.5s}&6& \textbf{0.5s} & 7 & \textbf{6.0s}& 8 &\textbf{6.0s} \\
& 4 & 1& \textbf{0.5s} & \textbf{0.5s} &  3& 3.0s & \textbf{1.0s} &  6& 141.0s & \textbf{3.5s} & 8&t/o & \textbf{4.0s}& 7 & 356.0s&\textbf{5.0s} & 8& \textbf{0.5s} &8& \textbf{5.0s}&8 &\textbf{5.0s}\\
\hline
\end{tabular}
}
\caption{PDDL Problem Instances}
\label{table1}
\end{table*}

\begin{table}[t]
  \centering
 \resizebox{1.0\columnwidth}{!} {
\begin{tabular}{|cc||cr|cr|cr|cr|} 
\hline
\multicolumn{2}{|c||}{\multirow{2}{*}{Prob.}} & \multicolumn{2}{c|}{Scenario 9}  & \multicolumn{2}{c|}{Scenario 10}  & \multicolumn{2}{c|}{Scenario 11}  & \multicolumn{2}{c|}{Scenario 12} \\
\multicolumn{2}{|c||}{} & $|\epsilon|$ & {\sc alg2}&\ $|\epsilon|$ & {\sc alg2} & $|\epsilon|$ & {\sc alg2} & $|\epsilon|$ & {\sc alg2}  \\
\hline
\hline
\multirow{3}{*}{\rotatebox{90}{\hspace{0.0em}\textsc{SAT}} \rotatebox{90}{\hspace{0.0em}\textsc{Grid}}} 
& 1   & 76 & 3.0s & 152&5.0s & 231&7.0s &298 &8.5s \\ 
& 2  &15 & 0.2s &17 &0.2s &19 &0.3s &20 &0.4s \\ 
& 3 & 177 &28.0s &354 &35.0s &531 &56.0s &708 &92.0s \\ 
\hline
\multirow{3}{*}{\rotatebox{90}{\hspace{0.0em}\textsc{BN}} \rotatebox{90}{\hspace{0.0em}\textsc{Ratio}}} 
& 1   &14 &0.2s &17 &0.2s & 19&0.3s &22 &0.5s \\ 
& 2  & 55 &1.0s &58 &2.0s & 59&2.0s &75 &6.0s \\ 
& 3   & 13& 6.0s& 26& 18.0s& 42& 47.0s& 51& 83.0s\\ 
\hline
\multirow{3}{*}{\rotatebox{90}{\hspace{0.0em}\textsc{Ace}} \rotatebox{90}{\hspace{0.0em}\textsc{}}} 
& 1  &13 &2.5s & 20 &12.0s &34& 46.0s& 45& 130.0s\\ 
& 2   &3 &1.5s &6 &9.0s &14 &55.0s & 16& 118.0s\\ 
& 3   &6 &8.5s &6 &37.0s &6 &131.0s &8 & 356.0s \\ 
\hline
\multirow{3}{*}{\rotatebox{90}{\hspace{0.0em}\textsc{BMC}} \rotatebox{90}{\hspace{0.0em}\textsc{}}} 
& 1   & 18 & 2.0s& 46& 12.0s & 69& 57.0s& 95& 193.0s \\ 
& 2   &6 & 8.0s &12 &43.0s &22 &398.0s & --& t/o \\ 
& 3  & 9 & 15.0s &9 &45.0s &20 &192.0s &-- &t/o \\ 
\hline
\end{tabular}
}
\caption{SAT Competition Problem Instances}
\label{table3}
\end{table}

We empirically evaluated Algorithm 2, referred to as {\sc alg2}, to find minimum explanations against the current planning-based state-of-the-art algorithm by \citet{chakraborti2017plan}, referred to as {\sc cszk}, which runs an A* search algorithm over the explanation search space and prioritizes actions that are in the plan being explained before others.\footnote{We used the implementation of the authors, which is publicly available at: \url{https://github.com/TathagataChakraborti/mmp}.}
We consider eight different ways to tweak the user's model, resulting in the following eight scenarios: 
\beitemize
\item \textbf{Scenario~1:}~We removed one random precondition from every action in the user's model. 
\item \textbf{Scenario~2:}~We removed one random effect from every action in the user's model. 
\item \textbf{Scenario~3:}~We removed one random precondition and one random effect from every action in the user's model. 
\item \textbf{Scenario~4:}~We removed multiple random preconditions and effects from every action in the user's model. 
\item \textbf{Scenario~5:}~We removed all preconditions from every action in the user's model. 
\item \textbf{Scenario~6:}~We removed multiple random predicates from the initial state in the user's model.
\item \textbf{Scenario~7:}~We removed all effects from every action in the user's model. 
\item \textbf{Scenario~8:}~We removed all actions in the user's model. 
\enitemize

Table~\ref{table1} tabulates the cardinality of the explanations $|\epsilon|$ as well as the runtimes of {\sc cszk} and {\sc alg2}. We did not report runtimes of {\sc cszk} for Scenario~6 as the available implementation was not designed to handle that scenario. We also omit the runtimes of {\sc cszk} for Scenarios~7 and~8 as the implementation contained errors on those instances that we were not able to repair, even though it theoretically could handle such scenarios. In general, {\sc alg2} is faster than {\sc cszk} in the majority of scenarios and domains; their runtimes for most problems in Scenarios~1 and~2 are similar, but {\sc alg2} is faster than {\sc cszk} in Scenarios~3 to~5 by up to two orders of magnitude. In general, the runtime of both algorithms increases as the difference between the models of the agent and user increases since both algorithms search over the explanation search space, which increases as the number of differences between the two models increases. 
However, {\sc alg2} is faster because it is able to use highly-optimized solvers to find MCSes, MUSes, and hitting sets.

\subsection{General Instances}
In order to test the generality and robustness of our proposed approach, we conducted experiments on a number of problem instances taken from recent SAT competitions. Similarly to the planning experiments, we used the actual instances as the model of the agent (i.e, $KB_{a}$), and tweaked that model and assigned it to be the model of the user (i.e.,~$KB_{h}$). The query that we used for each instance was a conjunction of backbone literals,\footnote{The backbone literals of a propositional $KB$ are the set of literals entailed by the $KB$.} which we computed using the \emph{minibones} algorithm proposed by~\citet{janota2015algorithms}.

 We consider four different ways to tweak the user's model, resulting in the following four scenarios: 
\beitemize
\item \textbf{Scenario~9:}~We randomly removed 10\% of the clauses and removed 20\% of literals from 10\% of the total clauses in the user's model.
\item \textbf{Scenario~10:}~We randomly removed 20\% of the clauses and removed 20\% of literals from 20\% of the total clauses in the user's model.
\item \textbf{Scenario~11:}~We randomly removed 30\% of the clauses and removed 20\% of literals from 30\% of the total clauses in the user's model.
\item \textbf{Scenario~12:}~We randomly removed 40\% of the clauses and removed 20\% of literals from 40\% of the total clauses in the user's model.

\enitemize

Table~\ref{table3} presents the results, where we report the explanation cardinality $|\epsilon|$ and runtimes of {\sc alg2}. We did not compare with any other system since, to the best of our knowledge, no such system exists. In general, {\sc alg2}, given the set time limit, managed to compute an explanation for all instances. In general, the runtimes of {\sc alg2} increase as the size of the knowledge bases and $|\epsilon|$ increase.

Moreover, similar to classical planning, we observe that the runtimes of {\sc alg2} increase as the difference between the two models increase (i.e.,~from Scenario~9 to Scenario~12). 

%% file: related.tex
\section{Related Work and Discussions}

Our work sits in the intersection of knowledge representation and planning -- our techniques are inspired by findings from the knowledge representation community, especially on MCS, MUS, and their duality; and our application problem of model reconciliation was introduced by the planning community in the context of explainable planning. Therefore, we will situate our work in the context of these two communities and relate to existing work in those two areas.

\smallskip \noindent \textbf{Related Work from Knowledge Representation:} 
The definition of the logic-based MRP we are tackling in this paper (see Definition~\ref{lmrp}) has been introduced in a previous work~\cite{vas}. Building on those theoretical foundations, in this paper, we consider the development of an efficient algorithm for computing minimal explanations, which further extends the logic-based MRP to account for knowledge bases that may become inconsistent after receiving an explanation.


The algorithm presented in this paper is inspired by a procedure
for computing an SMUS of an inconsistent formula, originally presented by~\citet{ignatiev-plm15}.
The method is also related to other similar approaches for enumerating MUSes and MCSes.
Moreover, our approach is similar in spirit to the HS-tree presented by~\citet{rei87}.
Although the original purpose was to enumerate diagnoses, Reiter's procedure can be
easily adapted to enumerate MUSes (called conflicts in that paper) as already noted by \citet{previti-m13}. 
However, the computation of an SMUS might require more substantial modifications. 
Procedures like the one presented by Reiter, which target MCSes (diagnoses)
instead of MUSes (conflicts), can be seen as the dual version of our algorithm. In particular,
the algorithm MaxHS~\cite{daviesB11} applies the same idea of iteratively computing and
testing a minimum hitting set for the computation of a MaxSAT solution (the complement
of the smallest MCSes).
%

On a different note, the notion of supports or explanations proposed in this paper and how they are used to update the knowledge base of the human might appear similar to the notion of belief revision~\cite{gardenfors1995belief}. However, there are some subtle differences. Belief revision usually refers to a single agent revising its belief after receiving a new piece of information that is in conflict with its current beliefs. As such, there is a temporal dimension in belief revision and a requirement that it should maintain as much as possible the belief of the agent, per AGM postulates \citep{alchourron1985logic,gardenfors1986belief}. On the other hand, our notion of explanation is done with respect to \emph{two knowledge bases} and there is no such requirement on maintaining as much as possible the belief of the human. Instead, our requirement is on minimizing the size of the explanation. 
Additionally, our notion of explanation might appear similar to the notion of a diagnosis (e.g.,~\cite{rei87}). Diagnosis focuses on identifying the reason for the inconsistency of a theory (i.e., $KB$) whereas an explanation aims at identifying the support for a formula. The difference lies in that a diagnosis is made with respect to the same theory (i.e., $KB_{a}$) while explanation is sought for the second theory (i.e., $KB_{h}$). For a further exposition on the relationship between our approach and previous works we refer the reader to~\cite{vasileiou2020relationship}.


\smallskip \noindent \textbf{Related Work from Planning:} 
Our work was motivated by the work of~\citet{chakraborti2017plan}, who introduced the model reconciliation problem that we are tackling in this paper. Hence, both approached share similarities, such as the types of explanations that can be found. For instance, the cardinality-minimal support in Definition 5 is equivalent to \emph{minimally complete explanations} (MCEs) (the shortest explanation). Our objective of finding a minimal explanation (actually, support) can be viewed as similar to the \emph{minimally monotonic explanations} (MMEs) (the shortest explanation such that no further model updates invalidate it). Similarly, \emph{model patch explanations} (MPEs) (includes all the model updates) are trivial explanations and are equivalent to our definition that $KB_a$ itself serves as an explanation for $KB_h$. Nonetheless, we would like to point out a fundamental difference between the two approaches: Our approach is based on KR (i.e., propositional logic in this paper), while theirs is based on automated planning and heuristic search techniques. Additionally, the explanation space search in our approach is done in the grounded representation of the planning domain, whereas theirs in the lifted representation. As a consequence, our approach can structurally isolate the reasons for a particular behavior. For example, an explanation from our approach could explain why a state (frame axioms) or an action (action dynamics) was true (or not) at a particular time step during the execution of the plan.

%% file: conclusions.tex
\section{Conclusions and Future Work}

In this paper, we presented a new logic-based approach that exploits the notion of hitting sets for model reconciliation problems.
The approach builds on top of previous techniques for computing SMUSes and it
shares with them the advantage of being constraint agnostic. Experiments showed that
our algorithm outperforms state-of-the-art alternatives in the context of planning and it is
able to solve representative SAT instances. Due to its logic-based nature, the approach
has the additional advantage of being able to deal with problems coming from different settings,
as far as the problem can be encoded into a constraint system. Here, we showed a propositional encoding
for the planning case. 

The algorithm we presented returns a smallest explanation $\epsilon$ such that when it is used to update the model of the user to $KB_h$, the updated model $KB_h \land \epsilon$ entails a formula $\varphi$ that needs to be explained. 
However, other kind of preferred explanations could be considered by defining a cost function over them. Future research will investigate
alternative preferred explanations. This should be possible by simply applying weights to the
elements in $\R$ and return a hitting set that minimizes the cost. Another line of research is to include all
the optimization techniques that have already been successfully applied for the extraction of an SMUS and consider the possibility of approximate solutions. Finally, we also plan to evaluate our approach with other encodings for other problems (e.g., SMT encodings for hybrid planning problems \cite{Cashmore_icaps2016}).

%% file: paper.bbl
\begin{thebibliography}{32}
\providecommand{\natexlab}[1]{#1}
\providecommand{\url}[1]{\texttt{#1}}
\providecommand{\urlprefix}{URL }
\expandafter\ifx\csname urlstyle\endcsname\relax
  \providecommand{\doi}[1]{doi:\discretionary{}{}{}#1}\else
  \providecommand{\doi}{doi:\discretionary{}{}{}\begingroup
  \urlstyle{rm}\Url}\fi

\bibitem[{Alchourr{\'o}n and Makinson(1985)}]{alchourron1985logic}
Alchourr{\'o}n, C.~E.; and Makinson, D. 1985.
\newblock On the logic of theory change: Safe contraction.
\newblock \emph{Studia logica} 44(4): 405--422.

\bibitem[{Alvarez~Melis and Jaakkola(2018)}]{alvarez2018towards}
Alvarez~Melis, D.; and Jaakkola, T. 2018.
\newblock Towards robust interpretability with self-explaining neural networks.
\newblock In \emph{NeurIPS}, 7775--7784.

\bibitem[{Biere et~al.(2009)Biere, Heule, van Maaren, and Walsh}]{sat-handbook}
Biere, A.; Heule, M.; van Maaren, H.; and Walsh, T., eds. 2009.
\newblock \emph{{Handbook of Satisfiability}}.

\bibitem[{Cashmore et~al.(2016)Cashmore, Fox, Long, and
  Magazzeni}]{Cashmore_icaps2016}
Cashmore, M.; Fox, M.; Long, D.; and Magazzeni, D. 2016.
\newblock {A Compilation of the Full {PDDL+} Language into {SMT}}.
\newblock In \emph{ICAPS}, 79–87.

\bibitem[{Chakraborti, Sreedharan, and Kambhampati(2019)}]{ijcai19}
Chakraborti, T.; Sreedharan, S.; and Kambhampati, S. 2019.
\newblock Balancing Explicability and Explanations in Human-Aware Planning.
\newblock In \emph{IJCAI}, 1335--1343.

\bibitem[{Chakraborti et~al.(2017)Chakraborti, Sreedharan, Zhang, and
  Kambhampati}]{chakraborti2017plan}
Chakraborti, T.; Sreedharan, S.; Zhang, Y.; and Kambhampati, S. 2017.
\newblock Plan Explanations as Model Reconciliation: Moving Beyond Explanation
  as Soliloquy.
\newblock In \emph{IJCAI}, 156--163.

\bibitem[{Davies and Bacchus(2011)}]{daviesB11}
Davies, J.; and Bacchus, F. 2011.
\newblock Solving {MAXSAT} by Solving a Sequence of Simpler {SAT} Instances.
\newblock In \emph{CP}, 225--239.

\bibitem[{Davis et~al.(1962)Davis, Logemann, Donald, and Loveland}]{DPLL62}
Davis, M.; Logemann, G.; Donald; and Loveland. 1962.
\newblock {A Machine Program for Theorem Proving}.
\newblock \emph{Communications of the ACM} 5(7): 394--397.

\bibitem[{de~Kleer, Mackworth, and Reiter(1992)}]{KleerMR92}
de~Kleer, J.; Mackworth, A.~K.; and Reiter, R. 1992.
\newblock Characterizing Diagnoses and Systems.
\newblock \emph{Artificial Intelligence} 56(2-3): 197--222.

\bibitem[{Dong et~al.(2017)Dong, Su, Zhu, and Zhang}]{dong2017improving}
Dong, Y.; Su, H.; Zhu, J.; and Zhang, B. 2017.
\newblock Improving interpretability of deep neural networks with semantic
  information.
\newblock In \emph{CVPR}, 4306--4314.

\bibitem[{G{\"a}rdenfors(1986)}]{gardenfors1986belief}
G{\"a}rdenfors, P. 1986.
\newblock Belief revisions and the Ramsey test for conditionals.
\newblock \emph{The Philosophical Review} 95(1): 81--93.

\bibitem[{G{\"a}rdenfors et~al.(1995)G{\"a}rdenfors, Rott, Gabbay, Hogger, and
  Robinson}]{gardenfors1995belief}
G{\"a}rdenfors, P.; Rott, H.; Gabbay, D.; Hogger, C.; and Robinson, J. 1995.
\newblock Belief Revision.
\newblock \emph{Computational Complexity} 63(6).

\bibitem[{Gilpin et~al.(2018)Gilpin, Bau, Yuan, Bajwa, Specter, and
  Kagal}]{gilpin2018explaining}
Gilpin, L.~H.; Bau, D.; Yuan, B.~Z.; Bajwa, A.; Specter, M.; and Kagal, L.
  2018.
\newblock Explaining Explanations: An Overview of Interpretability of Machine
  Learning.
\newblock In \emph{DSAA}, 80--89.

\bibitem[{Guidotti et~al.(2018)Guidotti, Monreale, Ruggieri, Turini, Giannotti,
  and Pedreschi}]{Guidotti:2018:SME:3271482.3236009}
Guidotti, R.; Monreale, A.; Ruggieri, S.; Turini, F.; Giannotti, F.; and
  Pedreschi, D. 2018.
\newblock A Survey of Methods for Explaining Black Box Models.
\newblock \emph{ACM Computing Survey} 51(5): 1--42.

\bibitem[{Ignatiev, Morgado, and Marques{-}Silva(2018)}]{imms-sat18}
Ignatiev, A.; Morgado, A.; and Marques{-}Silva, J. 2018.
\newblock {PySAT:} {A} {Python} Toolkit for Prototyping with {SAT} Oracles.
\newblock In \emph{SAT}, 428--437.

\bibitem[{Ignatiev et~al.(2015)Ignatiev, Previti, Liffiton, and
  Marques{-}Silva}]{ignatiev-plm15}
Ignatiev, A.; Previti, A.; Liffiton, M.~H.; and Marques{-}Silva, J. 2015.
\newblock Smallest {MUS} Extraction with Minimal Hitting Set Dualization.
\newblock In \emph{CP}, 173--182.

\bibitem[{Janota, Lynce, and Marques-Silva(2015)}]{janota2015algorithms}
Janota, M.; Lynce, I.; and Marques-Silva, J. 2015.
\newblock Algorithms for computing backbones of propositional formulae.
\newblock \emph{AI Communications} 28(2): 161--177.

\bibitem[{Kambhampati(1990)}]{kambhampati1990classification}
Kambhampati, S. 1990.
\newblock A classification of plan modification strategies based on coverage
  and information requirements.
\newblock \emph{AAAI Spring Symposium Series} .

\bibitem[{Kautz, McAllester, and Selman(1996)}]{kautz1996encoding}
Kautz, H.; McAllester, D.; and Selman, B. 1996.
\newblock Encoding plans in propositional logic.
\newblock In \emph{KR}, 374--384.

\bibitem[{Kautz and Selman(1992)}]{kautz1992planning}
Kautz, H.; and Selman, B. 1992.
\newblock Planning as Satisfiability.
\newblock In \emph{ECAI}, 359--363.

\bibitem[{Langley(2016)}]{langley2016explainable}
Langley, P. 2016.
\newblock Explainable agency in human-robot interaction.
\newblock \emph{AAAI Fall Symposium Series} .

\bibitem[{Li and Manya(2009)}]{li2009maxsat}
Li, C.~M.; and Manya, F. 2009.
\newblock MaxSAT, Hard and Soft Constraints.
\newblock \emph{Handbook of Satisfiability} 185: 613--631.

\bibitem[{Liffiton et~al.(2016)Liffiton, Previti, Malik, and
  Marques{-}Silva}]{liffiton-pmm16}
Liffiton, M.~H.; Previti, A.; Malik, A.; and Marques{-}Silva, J. 2016.
\newblock Fast, flexible {MUS} enumeration.
\newblock \emph{Constraints} 21(2): 223--250.

\bibitem[{Liffiton and Sakallah(2008)}]{kas-jar08}
Liffiton, M.~H.; and Sakallah, K.~A. 2008.
\newblock Algorithms for computing minimal unsatisfiable subsets of
  constraints.
\newblock \emph{Journal of Automated Reasoning} 40(1): 1--33.

\bibitem[{Marques{-}Silva et~al.(2013)Marques{-}Silva, Heras, Janota, Previti,
  and Belov}]{marques-silva-ijcai13}
Marques{-}Silva, J.; Heras, F.; Janota, M.; Previti, A.; and Belov, A. 2013.
\newblock On Computing Minimal Correction Subsets.
\newblock In \emph{IJCAI}, 615--622.

\bibitem[{Miller(2018)}]{miller2018explanation}
Miller, T. 2018.
\newblock Explanation in artificial intelligence: Insights from the social
  sciences.
\newblock \emph{Artificial Intelligence} 267: 1--38.

\bibitem[{Petkovic et~al.(2018)Petkovic, Altman, Wong, and
  Vigil}]{petkovic2018improving}
Petkovic, D.; Altman, R.; Wong, M.; and Vigil, A. 2018.
\newblock Improving the explainability of random forest classifier--user
  centered approach.
\newblock \emph{Pacific Symposium on Biocomputing} 23: 204--215.

\bibitem[{Previti and Marques{-}Silva(2013)}]{previti-m13}
Previti, A.; and Marques{-}Silva, J. 2013.
\newblock Partial {MUS} Enumeration.
\newblock In \emph{{AAAI}}, 818--825.

\bibitem[{Reiter(1987)}]{rei87}
Reiter, R. 1987.
\newblock A theory of diagnosis from first principles.
\newblock \emph{Artificial Intelligence} 32(1): 57--95.

\bibitem[{Vasileiou, Yeoh, and Son(2019)}]{vas}
Vasileiou, S.~L.; Yeoh, W.; and Son, T.~C. 2019.
\newblock A Preliminary Logic-based Approach for Explanation Generation.
\newblock In \emph{ICAPS Workshop on XAIP}, 132--140.

\bibitem[{Vasileiou, Yeoh, and Son(2020)}]{vasileiou2020relationship}
Vasileiou, S.~L.; Yeoh, W.; and Son, T.~C. 2020.
\newblock On the Relationship Between KR Approaches for Explainable Planning.
\newblock In \emph{ICAPS Workshop on XAIP}.

\bibitem[{Zhang et~al.(2017)Zhang, Sreedharan, Kulkarni, Chakraborti, Zhuo, and
  Kambhampati}]{zhang2017plan}
Zhang, Y.; Sreedharan, S.; Kulkarni, A.; Chakraborti, T.; Zhuo, H.~H.; and
  Kambhampati, S. 2017.
\newblock Plan explicability and predictability for robot task planning.
\newblock In \emph{ICRA}, 1313--1320.

\end{thebibliography}
